\documentclass[11pt,a4paper,usenames,table,dvipsnames]{article}
\usepackage[hyperref]{acl2020}
\usepackage{times}
\usepackage{latexsym}

\usepackage{amssymb} % allows symbols to be used 
    \usepackage{amsmath}
\usepackage{mathtools}
\usepackage{placeins}
\usepackage{bigfoot}
\usepackage{hyperref}
\usepackage{pbox}
\usepackage{svg}
\usepackage{todonotes}
\usepackage{tabularx}
\usepackage[linesnumbered,ruled,vlined]{algorithm2e}
\usepackage{amsfonts}
\usepackage{algpseudocode}
\usepackage{commath}
\usepackage{makecell}
\usepackage{adjustbox}

\DeclareMathOperator*{\argmax}{arg\,max}

\makeatletter
\def\thickhline{%
	\noalign{\ifnum0=`}\fi\hrule \@height \thickarrayrulewidth \futurelet
	\reserved@a\@xthickhline}
\def\@xthickhline{\ifx\reserved@a\thickhline
	\vskip\doublerulesep
	\vskip-\thickarrayrulewidth
	\fi
	\ifnum0=`{\fi}}
\makeatother

\newlength{\thickarrayrulewidth}
\usepackage{microtype}

\aclfinalcopy % Uncomment this line for the final submission
\usepackage{graphicx}

\title{Examining the State-of-the-Art in News Timeline Summarization}

\author{Demian Gholipour Ghalandari$^{1,2}$ \and Georgiana Ifrim$^1$ \\
$^1$Insight Centre for Data Analytics, University College Dublin, Ireland \\
$^2$Aylien Ltd., Dublin, Ireland \\
\texttt{\{demian.gholipour|georgiana.ifrim\}@insight-centre.org}} 

\date{}

\begin{document}
\maketitle
\begin{abstract}

%Automated generationcan reduce the effort of manually creating timelines and enables the coverage of more topics. Timeline summaries are a common solution to the problem of summarizing evolving news stories. 
Previous work on automatic news timeline
summarization (TLS) leaves an unclear picture about how this task can generally be approached and how well it is currently solved. This is mostly due to the focus on individual subtasks, such as date selection and date summarization, and to the previous lack of appropriate evaluation metrics for the full TLS task. In this paper, we compare different TLS strategies using appropriate evaluation frameworks, and propose a simple and effective combination of methods that improves over the  state-of-the-art on all tested benchmarks. For a more robust evaluation, we also present a new TLS dataset, which is larger and spans longer time periods than previous datasets. The dataset will be made available at \url{https://github.com/complementizer/news-tls}.
\end{abstract}

\section{Introduction}
Timelines of news events can be useful to condense long-ranging news topics and can help us understand how current major events follow from prior events.
Timeline summarization (TLS) aims to automatically create such timelines, i.e., temporally ordered time-stamped textual summaries of events focused on a particular topic. 
While TLS has been studied before, most works treat it as a combination of two individual subtasks, 1) date selection and 2) date summarization, and only focus on one of these at a time \cite{binh2013predicting, tran2013leveraging, tran2015joint}. However, these subtasks are almost never evaluated in combination, which leaves an unclear picture of how well TLS is being solved in general.
Furthermore, previously used evaluation metrics for the date selection and timeline summarization tasks are not appropriate since they do not consider the temporal alignment in the evaluation. Just until recently, there were no established experimental settings and appropriate metrics for the full TLS task \cite{martschat2017improving, martschat2018temporally}.

\begin{table}[!h]
\small
\centering
\begin{tabular}{|p{0.2\columnwidth}|p{0.67\columnwidth}|}
\hline
\textbf{Date} & \textbf{Summary} \\ \hline
2001-11-29 & Enron could cost Dutch group \$195m \\ \hline
2001-11-30 & 1,100 UK jobs go in Enron collapse \\ \hline
\cellcolor{blue!15} 2001-12-02 & \cellcolor{blue!15} Barclays: Enron bankruptcy will not affect business \\ \hline
\cellcolor{blue!15} 2002-01-15 & As Enron scandal spreads, US starts to question cash for influence culture\\ \hline
\multicolumn{2}{|l|}{{\hspace{3.4cm}[}...{]}} \\ \hline
2004-07-08 & \cellcolor{blue!15} Jury indicts Lay for inflating Enron earnings \\ \hline
\cellcolor{blue!15} 2006-05-25 & \cellcolor{blue!15} Former Enron bosses found guilty \\ \hline
\cellcolor{blue!15} 2006-07-05 & \cellcolor{blue!15} Enron founder Lay dies \\ \hline
2008-02-22 & US prison beckons British bankers who got cosy with Enron \\ \hline
\end{tabular}
\caption{Excerpt of an automatically constructed timeline about the company Enron, using article headlines as summaries. The shaded parts indicate that the date or summary matches entries in a corresponding human-written ground-truth timeline.}
\label{tab:intro-example}
\end{table}

In this paper, we examine existing strategies for the full TLS task and how well they actually work. We identify three high-level approaches: 1) \textit{Direct summarization} treats TLS like text summarization, e.g., by selecting a small subset of sentences from a massive collection of news articles; 2) The \textit{date-wise} approach first selects salient dates and then summarizes each date; 3) \textit{Event detection} first detects events, e.g., via clustering, selects salient events and summarizes these individually. The current state-of-the-art method is based on direct summarization \cite{martschat2018temporally}. We therefore focus on testing the  two remaining strategies, which have not been appropriately evaluated yet and allow for better scalability. 

We propose a simple method to improve date summarization for the date-wise approach. The method uses temporal expressions (textual references to dates) to derive date vectors, which in turn help to filter candidate sentences to summarize particular dates. With this modification, the date-wise approach obtains improved  state-of-the-art results on all tested datasets. We also propose an event-based approach via clustering, which outperforms  \cite{martschat2018temporally} on one of three tested datasets.
We use purpose-build evaluation metrics for evaluating timelines introduced by \citet{martschat2017improving}. For a more robust evaluation, we also present a new dataset for TLS, which is significantly larger than previous datasets in terms of the number of individual topics and time span. 
%(decades of news timelines).

% \begin{figure*}[]
%   \centering
%   \includegraphics[width=1\textwidth]{fig/phil_spector_timeline_crop.pdf}
%   \caption{Example of an automatically constructed timeline about Phil Spector.}
%   \label{fig:intro-example}
% \end{figure*}

We summarize our contributions as follows:
\begin{enumerate}
    \item We compare different TLS strategies side-by-side using suitable evaluation metrics to provide a better picture for how well the full TLS task for news is solved so far.
    \item We propose a simple addition to existing methods to significantly improve date-wise TLS, achieving new state-of-the-art results.
    
    \item We present a new TLS dataset that is larger than previous datasets and spans longer time ranges (decades of news timelines).
\end{enumerate}

\section{Related Work}

Timeline summarization for news articles has received some attention in the last two decades \cite{swan2000automatic, allan2001temporal, chieu2004query, yan2011timelineA, yan2011evolutionaryB, kessler2012finding, binh2013predicting, tran2013leveraging, li2013evolutionary, tran2015timeline, tran2015joint, wang2015socially, wang2016low, martschat2017improving, martschat2018temporally, steen2019abstractive}. The task is commonly split into date selection and date summarization subtasks. 

\subsection{Date Selection}
Supervised machine learning has been proposed to predict whether dates appear in ground-truth timelines \cite{kessler2012finding, binh2013predicting}. \citet{tran2015joint} use graph-based ranking of dates, which is reported to outperform supervised methods\footnote{Despite our best efforts, we could neither obtain code for this method from the authors nor reproduce its reported performance, and therefore did not include it in our experiments.}.

\subsection{Date Summarization}
Several approaches construct date summaries by picking sentences from ranked lists. The ranking is based on regression or learning-to-rank to predict ROUGE scores between the sentence and a ground-truth summary \cite{binh2013predicting, tran2013leveraging}. 
\citet{tran2015timeline} observe that users prefer summaries consisting of headlines to summaries consisting of sentences from article bodies. \citet{steen2019abstractive} propose abstractive date summarization based on graph-based sentence merging and compression.
Other works propose the use of additional data, such as comments on social media \cite{wang2015socially}, or images \cite{wang2016low}. 

\subsection{Full Timeline Summarization}
\citet{chieu2004query} produce timelines by ranking sentences from an entire document collection. The ranking is based on summed up similarities to other sentences in an $n$-day window.
\citet{nguyen2014ranking} propose a pipeline to generate timelines consisting of date selection, sentence clustering, and ranking.
\citet{martschat2018temporally} adapt submodular function optimization, commonly used for multi-document summarization, for the TLS task. The approach searches for a combination of sentences from a whole document collection to construct a timeline and is the current state-of-the-art for full TLS. \citet{steen2019abstractive} use a two-stage approach consisting of date selection and date summarization to build timelines.
 Other examples of automatic timeline generation can be found in the social media-related literature, where microblogs are often clustered before being summarized \cite{wang2014summarization, li2014timeline}. We explore a similar framework for evaluating clustering-based TLS.
 
\section{Strategies for Timeline Summarization}
\subsection*{Problem Definition}
\label{sec:problem}
We define the TLS setup and task as follows. Given is a set of news articles $A$, a set of query keyphrases $Q$, and a ground-truth (reference) timeline $r$, with $l$ dates that are associated with $k$ sentences on average, i.e., $m = k * l$ sentences in total. The task is to construct a (system) timeline $s$ that contains $m$ sentences, assigned to an arbitrary number of dates. A simpler and stricter setting can also be used, in which $s$ must contain exactly $l$ dates with $k$ sentences each.

\subsection*{Approach Types}
A number of different high-level approaches can be used to tackle this task:
\begin{enumerate}
    \item \textbf{Direct Summarization}: $A$ is treated as one set of sentences, from which a timeline is directly extracted, 
    e.g., by optimizing a sentence combination \cite{martschat2018temporally}, or by sentence ranking \cite{chieu2004query}. Among these, \citet{martschat2018temporally}'s solution for the full TLS task has state-of-the-art accuracy but does not scale well.
    \item \textbf{Date-wise Approach}:
    This approach selects $l$ dates and then constructs a text summary of $k$ sentences on average for each date. 
\item \textbf{Event Detection}:
    This approach first detects events in $A$, e.g., by clustering similar articles, and then identifies the $l$ most important events and summarizes these separately. 
\end{enumerate}
Since no prior work has analyzed the latter two categories for the full TLS task, we discuss and develop such approaches next.

\subsection{Date-wise Approach}
The approach described here mostly consists of existing building blocks, with a few but important modifications proposed from our side.

\subsubsection*{Defining the Set of Dates}
First, we identify the set of possible dates to include in a timeline. 
We obtain these from (i) the publication dates of all articles in $A$ and (ii) textual references of dates in sentences in $A$, such as 'last Monday', or '12 April'. 
We use the tool HeidelTime\footnote{\url{https://github.com/HeidelTime/heideltime}} \cite{StroetgenGertz2013} to detect and resolve textual mentions of dates.

\subsubsection*{Date Selection}
\begin{figure}[t]
  \centering
  \includegraphics[width=1\columnwidth]{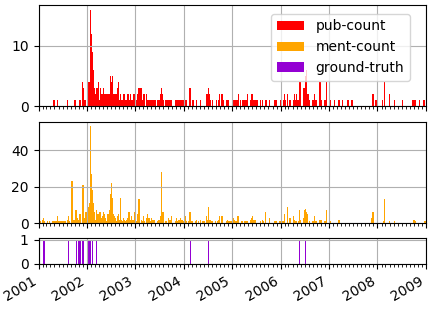}
  \caption{Counts of published articles and textual mentions across dates in an article collection about Enron.}
  \label{fig:frequencies}
\end{figure}
\label{sec:date-selection-approaches}
Next, we select the $l$ most important dates. We compare the following date selection methods introduced by \citet{binh2013predicting}:
\begin{itemize}
    \itemsep0em 
    \item \textsc{PubCount}: Ranking dates by the number of articles published on a date.
    \item \textsc{MentionCount}: Ranking dates by the number of sentences that mention the date.
    \item \textsc{Supervised}: Extracting date features and using classification or regression to predict whether a date appears in a ground-truth timeline. These features mostly include the publication count and different variants of counting date mentions.
\end{itemize}

Our experiments show that \textsc{Supervised} works best, closely followed by \textsc{MentionCount} (Appendix \ref{appendix:ds-variants}). Figure \ref{fig:frequencies} shows an example of publication and date mention counts and ground-truth dates over time. Two challenges are evident that date selection methods face: 1) These count signals usually do not perfectly correlate with ground-truth dates, and 2) high values often cluster around important dates, i.e., a "correct" date is often surrounded by other, "incorrect" dates with similarly strong signals. 

\subsubsection*{Candidate Sentences for Dates}

To summarize a particular date $d$, we first need to decide which articles or sentences we use as a source to create a summary from. Previous research has not explored this aspect much due to the separated treatment of subtasks. We propose a simple but effective heuristic to do this.
We consider the following two sets to be the primary source of suitable candidate sentences:
\begin{itemize}
    \itemsep0em 
    \item \textbf{$P_d$: Sentences \underline{published} on or closely after $d$}. These often contain initial reports of events occurring on $d$.
    \item \textbf{$M_d$: Sentences that \underline{mention} $d$}. These sentences are from articles published at any point in time, and may retrospectively refer to $d$, or announce events on $d$ beforehand\footnote{In practice, we include the first 5 sentences in the body of each article published on $d$ and up to 2 days after $d$ into $P_d$. We include all sentences found in $A$ that mention $d$ into $M_d$.}.
\end{itemize}

We evaluate these two options in our experiments, and propose an heuristic that combines these, which we call \textsc{PM-Mean}. We aim to find a subset of sentences in $P_d \cup M_d$ that are likely to mention important events happening on $d$. 
We convert all the sentences in the collection $A$ to sparse bag-of-words (unigram) vectors with sentence-level TF-IDF weighting.
We represent the sets of sentences $P_d$ and $M_d$ using the mean of their respective sentence vectors, $\overline{x}_{P_d}$ and $\overline{x}_{M_d}$.
The core assumption of the method is that the content shared between $P_d$ and $M_d$ is a good source for summarizing events on $d$. 
To capture this content, we build a \emph{date vector} $x_d$, so that we can compare sentence vectors against it to rank sentences. 
We set the value of $x_d$ for each dimension $i$ in the feature space as follows:

\begin{equation}
    \small
    {x}^{i}_{d} = \begin{cases}
    \frac{1}{|P_d|} \overline{x}^{i}_{P_d} + \frac{1}{|M_d|} \overline{x}^{i}_{M_d}
    & \text{if } \overline{x}^{i}_{P_d} > 0 \text{ and } \overline{x}^{i}_{M_d} > 0\\
    0
    & \text{otherwise}
    \end{cases}
\end{equation}

Thus the date vector $x_d$ is an average of $\overline{x}_{P_d}$ and $\overline{x}_{M_d}$ weighted by the sizes of $P_d$ and $M_d$, with any features zeroed out if they are missing in either $P_d$ or $M_d$.
To rank sentences, we compute the cosine similarity between the vector $x_s$ of each candidate sentence $s \in (P_d \cup M_d)$ to $x_d$. 
 We select the best-scoring candidate sentences by defining a threshold on this similarity. To avoid tuning this threshold, we use a simple knee point detection method
\cite{satopaa2011finding} to dynamically identify a threshold that represents the "knee" (or elbow) in the similarity distribution.
This set of best-scoring sentences is then used as the input for the final date summarization step.
%We draw a line from the minimum and maximum point of the similarity distribution and select the point with the maximum distance (orthogonal) to the line as the threshold.

\subsubsection*{Date Summaries}
To construct the final timeline, we separately construct a summary for the $l$ highest ranked dates.
Prior to our main experiments, we test several  multi-document summarization algorithms:
\begin{itemize}
    \itemsep0em 
    \item \textsc{TextRank}: Runs PageRank on a graph of pairwise sentences similarities to rank sentences \cite{mihalcea2004textrank}.
    \item \textsc{Centroid-Rank}: Ranks sentences by their similarity to the centroid of all sentences \cite{radev2004centroid}.
    \item \textsc{Centroid-Opt}: Greedily optimises a summary to be similar to the centroid of all sentences \cite{ghalandari2017revisiting}.
    \item \textsc{Submodular}: Greedily optimizes a summary using submodular objective functions that represent coverage and diversity \cite{lin2011class}. 
\end{itemize}
The only modification to these algorithms in our TLS pipeline is that we prevent sentences not containing any topic keyphrases from query $Q$ to be included in the summary. \textsc{Centroid-Opt} has the best results (Appendix \ref{appendix:ds-variants}) and is used in the main experiments. 

\subsubsection*{Timeline Construction}
The date-wise approach constructs a timeline as follows: first, rank all potential dates using one of the date selection approaches described, then pick the $l$ highest ranked ones, pick candidate sentences for each date, and summarize each date individually from the according candidate set, using $k$ sentences.
We might not be able to summarize a particular date due to the keyword constraint in the summarization step. Whenever this is the case, we skip to the next date in the ranked list, until $l$ is reached.

\subsection{Event Detection Approach}
\label{sec:clustering}

When humans are tasked with constructing a timeline, we expect that they reason over important \textit{events} rather than dates. Conceptually, detecting and selecting events might also be more appropriate than selecting dates because multiple events can happen on the same day, and an event can potentially span multiple days.

To explore this, we test a TLS approach based on event detection by means of article clustering. The general approach can be summarized as follows:
(1) Group articles into clusters; (2) Rank and select the $l$ most important clusters; (3) Construct a summary for each cluster. Similarly to the date-wise approach, this mostly consists of existing building blocks that we adapt for TLS.

%\subsubsection{Article Preprocessing}
\subsubsection*{Clustering}
For each input collection $A$, we compute sparse TF-IDF unigram bag-of-words vectors for all articles in $A$. We apply clustering algorithms to these vectors.
To cluster articles, we use Markov Clustering (MCL) with a temporal constraint.
MCL \cite{van2000graph} is a clustering algorithm for graphs, i.e., a community detection algorithm. It is based on simulating random walks along nodes in a graph. \citet{ribeiro2017unsupervised} use this approach for clustering news articles. 

We convert $A$ into a graph where nodes correspond to articles so that we can cluster the articles using MCL, with the following temporal constraint: Articles $a_1$, $a_2$ are assigned an edge if their publication dates are at most 1 day apart from each other. The edge weight is set to the cosine similarity between the TF-IDF bag-of-words vectors of $a_1$ and $a_2$.
The constraint on the publication dates ensures that clusters do not have temporal gaps. Furthermore, it reduces the number of similarity computations between pairs of articles considerably. We run MCL on this graph and obtain clusters by identifying the connected components in the resulting connectivity matrix\footnote{We use the implementation and default parameters from \url{https://github.com/GuyAllard/markov_clustering}}.

\subsubsection*{Assigning Dates to Clusters}
We define the cluster date as the date that is most frequently mentioned within articles of the cluster. We identify date mentions using the HeidelTime tool.

\subsubsection*{Cluster Ranking}
To construct a timeline, we only need the $l$ most important clusters. We obtain these by ranking and retaining the top-$l$ clusters of the ranked list. 
We test the following scores to rank clusters by:
\begin{itemize}
    \itemsep0em 
    \item \textsc{Size}: Rank by the numbers of articles in a cluster.
    \item \textsc{DateMentionCount}: Rank by how often the cluster date is mentioned throughout the input collection.
    \item \textsc{Regression}: Rank using a score by a regression model trained to predict importance scores of clusters.
\end{itemize}

For the regression-based ranking method, we represent clusters using the following features:
%\begin{itemize}
 number of articles in a cluster; number of days between the publication dates of the first and last article in the cluster; maximum count of publication dates of articles within a cluster; maximum mention count of dates mentioned in articles in a cluster; sum of mention counts of dates mentioned in articles in a cluster.
%\end{itemize}
We test two approaches to label clusters with target scores to predict. 
\begin{itemize}
    \itemsep0em 
    \item Date-Accuracy: This is 1 if the cluster date appears in the ground-truth, else 0.
    \item ROUGE: The ROUGE-1 F1-score\footnote{ROUGE-1 obtained a better overall performance than ROUGE-2 for this purpose.} between the summary of the cluster and the ground-truth summary of the cluster date. If the cluster date does not appear in the ground-truth, the score is set to 0.
\end{itemize}

We evaluate these different options  (Appendix \ref{appendix:clust-variants}) and observe that ranking by \textsc{DateMentionCount} works better than the supervised methods, showing that predicting the suitability of clusters for timelines is difficult.

\subsubsection*{Cluster Summarization}
We use the same multi-document summarization method that works best for the date-wise approach (\textsc{Centroid-Opt}).

\subsubsection*{Timeline Construction}
In summary, the clustering approach builds a timeline as follows: 1) cluster all articles, 2) rank clusters, 3) build a summary with $k$ sentences for the top-$l$ clusters, skipping clusters if a summary cannot be constructed due to missing keywords. Furthermore, we skip clusters if the date assigned to the cluster is already "used" by a previously picked cluster. Conceptually, this implies that we can only recognize one event per day. In initial experiments, this leads to better results than alternatives, e.g., allowing multiple summaries of length $k$ per day.

\section{Dataset}

Tran et al. introduced the 17 Timelines (\textsc{T17}) \cite{binh2013predicting} and the \textsc{Crisis} \cite{tran2015timeline} datasets for timeline summarization from news articles.
However, we see the need for better benchmarks due to 1) a small number of topics in the \textsc{T17} and \textsc{Crisis} datasets (9 and 4 topics respectively), and 2) relatively short time span, ranging from a few months to 2 years.

Therefore, we build a new TLS dataset, called \textsc{Entities}, that contains more topics (47) and longer time-ranges per topic, e.g., decades of news articles. 
In the following, we describe how we obtain ground-truth timelines and input article collections for this dataset.

\textbf{Ground-Truth Timelines}: We obtain ground-truth timelines from CNN Fast Facts\footnote{\url{http://edition.cnn.com/specials/world/fast-facts}}, which has a collection of several hundred timelines grouped in categories, e.g., `people' or `disasters'. We pick all timelines of the `people' category and a small number from other categories.

\textbf{Queries}: For each ground-truth timeline, we define a set of query keyphrases $Q$. By default, we use the original title of the timeline as the keyphrase. For people entities, we use the last token of the title to capture surnames only, which increases the coverage. We manually inspect the resulting sets of keyphrases and correct these if necessary.

\textbf{Input Articles}:
For each entity from the ground-truth timelines, we search for news articles using TheGuardian API\footnote{\url{http://open-platform.theguardian.com/}}. We use this source because it provides access to all published articles starting from 1999. We search for articles that have exact matches of the queries in the article body. The timespan for the article search is set so that it extends the ground-truth timeline by 10\% of its days before its first and after its last date.

\textbf{Adjustments and Filtering}: 
The ground-truth timelines are modified to be usable for TLS and to ensure they do not contain data not present in the document collection:
\begin{itemize}
    \setlength\itemsep{-0.2em}
    \item We remove entries in the ground-truth timelines if they do not specify year, month, and day of an event.
    \item Ground-truth timelines are truncated to the first and last date of the input articles.
    \item Entries in the ground-truth timeline are removed if there is no input article published within $\pm$ 2 days.
\end{itemize}
Afterwards, we remove all topics from the dataset that do not fulfill the following criteria:
\begin{itemize}
    \setlength\itemsep{-0.2em}
    \item The timeline must have at least 5 entries.
    \item For at least 50\% of the dates present in the ground-truth timeline, textual references have to be found in the article collection (e.g., 'on Wednesday' or 'on 1 August'.). This is done to ensure that the content of the timelines is reflected to some degree in the article collection.
    \item There are at least 100 and less than 3000 articles containing the timeline-entity in the input articles. This is done to reduce the running time of experiments.
\end{itemize}

\textbf{Dataset Characteristics}: 
Tables \ref{tab:dataset-stats-1} and \ref{tab:dataset-stats-2} give an overview of properties of the two existing datasets and our new dataset, and mostly show averaged values over tasks in a dataset. An individual task corresponds to one ground-truth timeline that a TLS algorithm aims to simulate. $\#PubDates$ refers to the number of days in an article collection $A$ on which any articles are published. The compression ratio w.r.t. sentences ("comp. ratio (sents)") is $m$ divided by the total number of sentences in $A$, and the compression ratio w.r.t dates is $l$ divided by $\#PubDates$. "Avg. date cov" refers to the average coverage of dates in the ground-truth timeline $r$ by the articles in $A$. This can be counted by using publication dates in $A$, ("published"), or by textual date references to dates within articles in $A$ ("mentioned"). The fact that there are generally more ground-truth dates covered in textual date references compared to publication dates suggests making use of these date mentions. 

\textsc{T17} has longer ($l$), and more detailed ($k$) timelines than the other datasets, \textsc{Crisis} has more articles per task, and \textsc{Entities} has more topics, publication dates and longer time periods per task.

\begin{table*}[!htb]
    \small
    \centering
    \caption{Dataset Statistics for the TLS task (i)}
    \begin{tabular}{|l|l|l|l|l|l|l|}
    \hline
    \textbf{Dataset} &
    \#Topics &
    \#TLs &
    \makecell{Avg. \\ \#Docs} & 
    \makecell{Avg. \\ \#Sents} & 
    \makecell{Avg. \\ \#PubDates} & 
    \makecell{Avg. \\ Duration \\ (in days)} \\ \hline
    \textsc{T17} & 9 & 19 & 508 & 20409 & 124 & 212 \\ 
    \textsc{Crisis} & 4 & 22 & 2310 & 82761 & 307 & 343 \\ 
    \textsc{Entities} & 47 & 47 & 959 & 31545 & 600 & 4437 ($\approx 12$ years) \\ \hline
    \end{tabular}
        \label{tab:dataset-stats-1}
\end{table*}

\begin{table*}[h]
    \small
    \centering
        \caption{Dataset Statistics for the TLS task (ii)}
    \begin{tabular}{|l|l|l|l|l|l|l|l|}
        \hline
        
        \textbf{Dataset} &
        \makecell{Avg. \\ l} & 
        \makecell{Avg. \\ k} & 
        \makecell{Avg. \\ m} & 
        \makecell{Avg. comp. \\ ratio (sents)} &
        \makecell{Avg. comp. \\ ratio (dates)} &
        \makecell{Avg. date cov. \\ (published)} &
        \makecell{Avg. date cov. \\ (mentioned)} \\ \hline
        \textsc{T17} & 36 & 2.9 & 108 & 0.0117 & 0.43 & 81\% & 93\% \\ 
        \textsc{Crisis} & 29 & 1.3 & 38 & 0.0005 & 0.11 & 90\% & 96\% \\ 
        \textsc{Entities} & 23 & 1.2 & 26 &  0.0017 & 0.06 & 51\% & 65\% \\ \hline
    \end{tabular}

    \label{tab:dataset-stats-2}
\end{table*}

\section{Experiments}
\subsection{Evaluation Metrics}
In our experiments, we measure the quality of generated timelines with the following two evaluation metrics, which are also used by \citet{martschat2018temporally}:
\begin{itemize}
    \itemsep0em
    \item \textbf{Alignment-based ROUGE F1-score:} This metric compares the textual overlap between a system and a ground-truth timeline, while also considering the assignments of dates to texts.
    \item \textbf{Date F1-score:} This metric compares only the dates of a system and a ground-truth timeline.
\end{itemize} 
We denote the alignment-based ROUGE-1 F1-score as AR1-F and Date F1-score as Date-F1.
\subsection{Experimental Settings}
Concerning the datasets and task, we follow the experimental settings of \citet{martschat2018temporally}:
\begin{itemize}
    \itemsep0em 
    \item Each dataset is divided into multiple topics, each having at least one ground-truth timeline. If a topic has multiple ground-truth timelines, we split the topic into multiple tasks. The final results in the evaluation are based on averages over tasks/ground-truth timelines, not over topics.
    \item Each task includes a set of news articles $A$, a set of keyphrases $Q$, a ground-truth timeline $r$, with number of dates (length) $l$, average number of summary sentences per date $k$, and total number of summary sentences $m = l * k$. 
    \item In each task, we remove all articles from $A$ whose publication dates are outside of the range of dates of the ground-truth timeline $r$ of the task. Article headlines are not used.
    \item We run leave-one-out cross-validation over all tasks of a dataset.
    \item We test for significant differences using
    an approximate randomization test \cite{marcus1993building} 
    with a p-value of 0.05. 
\end{itemize}

We use the following configurations for our methods:
\begin{itemize}
\itemsep0em
\item A stricter and simpler version of the output size constraint: We produce timelines with the number of dates $l$ and $k$ sentences per date.
\item  In the summarization step of our methods, we only allow a sentence to be part of a summary if it contains any keyphrase in $Q$. As opposed to \citet{martschat2018temporally}, we still keep sentences not matching $Q$, e.g., for TF-IDF computation, clustering, and computing date vectors.
\end{itemize}

\subsection{Methods Evaluated}

%\subsubsection{Tested Methods}
We compare the following types of methods to address the full news TLS task.

\textbf{Direct summarization approaches:}
\begin{itemize}
    \itemsep0em 
    \item \textsc{Chieu2004}: \citet{chieu2004query} An unsupervised baseline based on direct summarization. We use the reimplementation from \citet{martschat2018temporally}.
    \item \textsc{Martschat2018}: \citet{martschat2018temporally} State-of-the-art method on the \textsc{Crisis} and \textsc{T17} datasets. It greedily selects a combination of sentences from the entire collection $A$ maximizing submodular functions for content coverage, textual and temporal diversity, and a high count of date references\footnote{Multiple variants of this approach were introduced in the paper. We picked the variant called "AsMDS+$f_{\textrm{TempDiv}}$+$f_{\textrm{DateRef}}$" due to its good results.}.
\end{itemize}
\textbf{Date-wise approaches:}
\begin{itemize}
    \itemsep0em 
    \item \textsc{Tran 2013} \cite{binh2013predicting}: The original date-wise approach, using regression for both date selection and summarization, and using all sentences of a date as candidate sentences.
    \item \textsc{PubCount}: A simple date-wise baseline that uses the publication count to rank dates, and all sentences published on a date for candidate selection.  We use \textsc{Centroid-Opt} for summarization.
    \item \textsc{Datewise}: Our date-wise approach after testing different building blocks (see Appendix \ref{appendix:ds-variants}). It uses supervised date selection, \textsc{PM-mean} for candidate selection and  \textsc{Centroid-Opt} for summarization.
\end{itemize}
\textbf{Event detection approach based on clustering:}
\begin{itemize}
    \item \textsc{Clust}: We use \textsc{DateMentionCount} to rank clusters, and
    \textsc{Centroid-Opt} for summarization, which are the best options according to our tests (see Appendix \ref{appendix:clust-variants}).
\end{itemize}

Note that all methods apart from \textsc{Datewise} and \textsc{Clust} have been proposed previously.

\subsubsection*{Oracles:}
To interpret the alignment-based ROUGE scores better and to approximate their upper bounds, we measure the performance of three different oracle methods:
\begin{itemize}
    \itemsep0em 
    \item \textsc{Date Oracle}: Selects the correct (ground-truth) dates and uses \textsc{Centroid-Opt} for date summarization.
    \item \textsc{Text Oracle}: Uses regression to select dates, and then constructs a summary for each date by optimizing the ROUGE to the ground-truth summaries.
    \item \textsc{Full Oracle}: Selects the correct dates and constructs a summary for each date by optimizing the ROUGE to the ground-truth summaries.
\end{itemize}
We give more detail about these in Appendix \ref{appendix:oracles}.

\subsection{Results}

Table \ref{table:final-results} shows the final evaluation results. 
We reproduced the results of \textsc{Chieu2004} and \textsc{Martschat2018} reported by \citet{martschat2018temporally} using their provided code\footnote{With the exception of \textsc{Crisis} due to memory issues.}. The other results are based on our implementations. Table \ref{tab:examples} in Appendix \ref{appendix:examples} shows several output examples across different methods. 

\begin{table*}[h!]
    \resizebox{\textwidth}{!}{%
    \begin{tabular}{|l|lll|}
        \hline
        & \multicolumn{3}{c|}{\textsc{T17} Dataset} \\
        & \textbf{AR1-F} & \textbf{AR2-F} & \textbf{Date-F1} \\ \hline
        \textbf{Text Oracle} & 0.198 & 0.073 & 0.541 \\
        \textbf{Date Oracle} & 0.179 & 0.057 & 0.926 \\
        \textbf{Full Oracle} & 0.312 & 0.128 & 0.926 \\ \hline
        \textbf{\textsc{Chieu2004}} & 0.066 & 0.019 & 0.251 \\
        \textbf{\textsc{Martschat2018}} & 0.105 & 0.03 & \textbf{0.544} $\bullet$ \\        
        \textbf{\textsc{Tran2013}} & 0.094 & 0.022 & 0.517 $\bullet$ \\
        \textbf{\textsc{PubCount}} & 0.105 & 0.027 & 0.481 \\
        \textbf{\textsc{Datewise}} & \textbf{0.12} $\star \bullet \dagger$& \textbf{0.035} $\star \bullet$ & \textbf{0.544} $\star \bullet$\\
        \textbf{\textsc{Clust}} & 0.082 & 0.020 & 0.407 \\ \hline
        \textbf{\textsc{Datewise} (titles)} & - & - & - \\
        \hline
        \end{tabular}

    \begin{tabular}{|lll|}
    \hline
    \multicolumn{3}{|c|}{\textsc{Crisis} Dataset} \\
    \textbf{AR1-F} & \textbf{AR2-F} & \textbf{Date-F1} \\ \hline
    0.136 & 0.052 & 0.297 \\
    0.202 & 0.063 & 0.974 \\
    0.367 & 0.15 & 0.974 \\ \hline
    0.052 & 0.012 & 0.142 \\
    0.075 $\bullet$ & 0.016 & 0.281\\        
    0.054 & 0.011 & 0.289 \\
    0.067 & 0.012 & 0.233 \\
    \textbf{0.089} $\star \bullet$ & \textbf{0.026} $\star \bullet$ & \textbf{0.295} $\bullet$ \\
    0.061 & 0.013 & 0.226 \\ \hline
    0.072 & 0.016 & 0.287 \\ \hline
    \end{tabular}

    \begin{tabular}{|lll|}
    \hline
    \multicolumn{3}{|c|}{\textsc{Entities} Dataset} \\
    \textbf{AR1-F} & \textbf{AR2-F} & \textbf{Date-F1} \\ \hline
     0.069 & 0.023 & 0.20 \\
     0.17 & 0.047 & 0.757 \\
     0.232 & 0.075 & 0.757 \\ \hline
     0.036 & 0.01 & 0.102 \\
     0.042 & 0.009 & 0.167 \\        
     0.042 & 0.012 & 0.184 $\dagger$ \\
     0.033 & 0.009 & 0.107 \\
    \textbf{0.057} $\star \bullet \dagger$ & \textbf{0.017}$\star \bullet \dagger$ & \textbf{0.205} $\star \bullet \dagger$ \\
     0.051 $\dagger$ & 0.015 $\dagger$ & 0.174 \\ \hline
    0.057 & 0.017 & 0.194 \\ \hline
    \end{tabular}
    }
    \\
\caption{Results on the full TLS task. $\star$ indicates a significant improvement over Tran 2013, $\bullet$ over \textsc{Clust}, and $\dagger$ over \textsc{Martschat2018}. \textsc{Datewise} (titles) is not included in the significance testing.}
\label{table:final-results}    
\end{table*}

\section{Analysis and Discussion}

\subsection{Performance of TLS Strategies}

Among the methods evaluated, \textsc{Datewise} consistently outperforms all other methods on all tested datasets in the alignment-based ROUGE metrics. The Date-F1 metric for this method is close to other methods, and not always better, which shows that the advantage of \textsc{Datewise} is due to the sentence selection (based on our heuristic date vectors) and summarization. Note that the date selection method is identical to \textsc{Tran2013}. We conclude from these results that the expensive combinatorial optimization used in \textsc{Martschat2018} is not necessary to achieve high accuracy for news TLS.

\textsc{Clust} performs worse than \textsc{Datewise} and  \textsc{Martschat2018}, except on \textsc{Entities}, where it outperforms \textsc{Martschat2018}. We find that for the other two datasets, \textsc{Clust} often merges articles from close dates together that would belong to separate events on ground-truth timelines, which may suggest that a different granularity of clusters is required depending on the task.

\textsc{Date Oracle} and \textsc{Full Oracle} should theoretically have a 100\% Date-F1. In practice, their Date-F1 scores turn out lower because, for some dates, no candidate sentences that match query $Q$ can be found, which causes the dates to be omitted from the oracle timelines.

Based on the performance of different systems, the hardest dataset is \textsc{Entities}, followed by \textsc{Crisis}. 

\subsection{What makes TLS difficult?}
While the ranking of methods is fairly stable, the performance of all methods varies a lot across the datasets and across individual tasks within datasets. To find out what makes individual tasks difficult, we measure the Spearman correlation between AR1-F and several dataset statistics. The details are included in Appendix \ref{appendix:correlations}. The correlations show that a high number of articles and publication dates and a low compression ratio w.r.t to dates generally decreases performance. This implies that highly popular topics are harder to summarize. The duration of a topic also corresponds to lower performance, but in a less consistent pattern.

The generally low performance across tasks and methods is likely influenced by the following factors:
\begin{itemize}
    \item The decision for human editors to include particular events in a timeline and to summarise these in a particular way can be highly subjective. Due to the two-stage nature of TLS, this problem is amplified in comparison to regular text summarization.
    \item Article collections can be insufficient to cover every important event of a topic, e.g., due to the specific set of news sources or the search technique used.
\end{itemize}

\subsection{Running Time}
\textsc{Datewise} and \textsc{Clust} are up to an order of magnitude faster to run than \textsc{Martschat2018} (Appendix \ref{appendix:running-time}) since their date summarization steps only involve a small subset of sentences in an article collection.

\subsection{Adjacent Dates and Redundancy}
Automatically constructed timelines often contain a high amount of multiple adjacent dates, while this is not the case in ground-truth timelines. Summaries of such adjacent dates often tend to refer to the same event and introduce redundancy into a timeline. To quantify this, we count the proportion of those "date bigrams" in a chronologically ordered timeline, which are only 1 day apart. The results (see Table \ref{table:adjacent-dates}) show that this is an issue for \textsc{Martschat2018} and \textsc{Datewise}, but less so for \textsc{Clust}, which is designed to avoid this behavior. Note that \textsc{Martschat2018} includes an objective function to reward diversity within a timeline, while \textsc{Datewise} has no explicit mechanism against redundancy among separate dates. Interestingly, when forcing \textsc{Datewise} to avoid selecting adjacent dates (by skipping such dates in the ranked list), the performance in all metrics decreases. In this case, high redundancy is a safer strategy for optimizing TLS metrics compared to enforcing a more balanced spread over time. Because of such effects, we advise to use automated evaluation metrics for TLS with care and to conduct qualitative analysis and user studies where possible.

\subsection{Use of Titles}
While using article titles can make timelines more readable and understandable \cite{tran2015timeline}, we do not involve titles in our main experiments, in order to directly compare to 
\textsc{Martschat2018}, and due to the lack of titles in \textsc{T17}. The last row in Table \ref{table:final-results} shows the results of a separate experiment with \textsc{Datewise} in which we build date summaries using titles only. Using only titles generally increases AR Precision at the cost of Recall. AR-F is negatively affected in \textsc{Crisis} but does not change in \textsc{Entities}. Figure \ref{tab:intro-example} shows parts of a title-based timeline produced by \textsc{Datewise}.

\begin{table}[]
\resizebox{\columnwidth}{!}{%
\begin{tabular}{|l|llll|}
\hline
 & Ground-truth & \textsc{Martschat2018} & \textsc{Datewise} & \textsc{Clust}  \\\hline
\textsc{T17} & 0.45 & 0.63 & 0.62 & 0.25 \\
\textsc{Crisis} & 0.18 & - & 0.52 & 0.06 \\
\textsc{Entities} & 0.03 & 0.18 & 0.3 & 0.05 \\ \hline
\end{tabular}%
}
\caption{Proportion of adjacent dates of timelines produced by different methods, and the ground-truth timelines.}
\label{table:adjacent-dates}
\end{table}

\section{Conclusion}

In this study, we have compared and proposed different strategies to construct timeline summaries of long-ranging news topics: the previous state-of-the-art method based on direct summarization, a date-wise approach, and a clustering-based approach. By exploiting temporal expressions, we have improved the date-wise approach and yielded new state-of-the-art results on all tested datasets. Hence, we showed that an expensive combinatorial search over all sentences in a document collection is not necessary to achieve good results for news TLS.
For a more robust and diverse evaluation, we have constructed a new TLS dataset with a much larger number of topics and with longer time-spans than in previous datasets. 
Most of the generated timelines are still far from oracle timeline extractors and leave large gaps for improvements. 
Potential future directions include a more principled use of our proposed heuristic for detecting content relevant to specific dates, the use of abstractive techniques, a more effective treatment of the redundancy challenge, and extending the new dataset with multiple sources.

\section*{Acknowledgments}
This work was funded by the Irish Research Council (IRC) under grant number EBPPG/2018/23, the Science Foundation Ireland (SFI) under grant number 12/RC/2289\_P2 and the enterprise partner Aylien Ltd. 

\bibliographystyle{ACM-Reference-Format}
\bibliography{references}

\appendix
\section{Appendices}

\subsection{Testing Variants of \textsc{Datewise}}
\label{appendix:ds-variants}

Table \ref{table:ds-variants} shows results for different variants of the date-wise approach.

\textbf{Date Selection}: While testing different date selection methods, we use \textsc{PM-mean} for candidate selection and \textsc{Centroid-Opt} for summarization. The supervised date selection methods work best, closely followed by \textsc{MentionCount}. 

\textbf{Candidate Sentence Selection}: We compare different strategies of defining the set of sentences associated with a date prior to summarization. The results show that the PM-method can improve the performance, especially for the Crisis dataset.

\textbf{Date Summarization}: Finally, we test different unsupervised text summarization algorithms to summarize each selected date. \textsc{Centroid-Opt} works best and is used in our main experiments.

\begin{table}[htb!]
    \small
    \centering
    \begin{tabular}{|l|lll|}
        \hline
        \textbf{Date Selector} & T17 & Crisis & Entities \\ \hline
        PubCount & 0.49 & 0.243 & 0.135 \\
        MentCount & 0.528 & 0.295 & 0.159 \\
        Tran 2013 (Reg) & 0.535 & \textbf{0.297} & \textbf{0.191} \\
        Tran 2013 (Clf) & \textbf{0.541} & 0.295 & 0.172 \\ 
        \hline
        \hline
        \textbf{Candidate Selector} & T17 & Crisis & Entities \\ \hline
        Sents mentioning d & 0.11 & 0.077 & 0.041 \\ \hline
        \makecell{Sents published on \\ d to d + 2 (first 5)} & 0.112 & 0.078 & 0.045 \\ \hline
        \makecell{Sents published on \\ d to d + 2 (all)} & 0.113 & 0.079 & 0.041 \\ \hline
        \textsc{PM-Mean} & \textbf{0.118} & \textbf{0.089} & \textbf{0.047} \\
        \hline
        \hline
        \textbf{Summarizer} & T17 & Crisis & Entities \\ \hline
        \textsc{TextRank} & 0.113 & 0.086 & 0.046 \\
        \textsc{Centroid-Rank} & 0.112 & 0.085 & 0.046 \\
        \textsc{Centroid-Opt} & \textbf{0.118} & \textbf{0.089} & \textbf{0.047} \\
        \textsc{Submodular} & 0.116 & 0.088 & \textbf{0.047} \\ \hline
    \end{tabular}
\caption{Variants of Date-wise TLS and their alignment-based ROUGE-1 score.}
\label{table:ds-variants}
\end{table}

\subsection{Testing Variants of \textsc{Clust}}
\label{appendix:clust-variants}

For the clustering-based TLS approach, we only test different options for ranking clusters. For the summarization step, we use \textsc{Centroid-Opt}, which works best for the date-wise approach. Table \ref{table:clust-variants} shows somewhat inconsistent results, but overall \textsc{DateMentionCount} obtains the best performance in terms of alignment-based ROUGE.

\begin{table}[h]
    \small
    \setlength{\tabcolsep}{5pt}
    \begin{tabular}{|l|lll|}
    \hline
    \textbf{\makecell{Cluster Ranking (AR1-F)}} & T17 & Crisis & Entities \\ \hline
    \textsc{Size} & 0.08 & 0.06 & 0.048 \\
    \textsc{DateMentionCount} & 0.081 & \textbf{0.061} & \textbf{0.051} \\
    \textsc{Regression (Dates)} & 0.08 & 0.055 & 0.045 \\
    \textsc{Regression (Rouge)} & \textbf{0.082} &  0.055 & 0.048 \\
    \hline
    \textbf{\makecell{Cluster Ranking (Date-F1)}} & T17 & Crisis & Entities \\ \hline
    \textsc{Size} & 0.41 & 0.22 & \textbf{0.16}\\
    \textsc{DateMentionCount} & 0.41 & 0.23 & 0.15 \\
    \textsc{Regression (Dates)} & \textbf{0.46} & 0.23 & \textbf{0.16} \\
    \textsc{Regression (Rouge)} & 0.44 & \textbf{0.24} & \textbf{0.16} \\
    \hline
    \end{tabular}
    \caption{Variants of Date Clustering-based TLS.} 
    \label{table:clust-variants}
\end{table}

\subsection{Oracles}
\label{appendix:oracles}

For the text and full oracles, we use Algorithm \ref{alg:oracle} for constructing a summary for a date, using ROUGE-1 F1-score as the objective. 
We include all sentences that mention $d$, as well as the first 5 sentences of all articles published between $d$ and $d + 5$ days, as candidate sentences for the oracles to summarize dates.

\begin{algorithm} 
    \small
    {\bf Input}: Candidate sentences $C$, reference summary $R$, summary length $k$
    
    {\bf Output}: Summary sentences $S$
    
    {$S \leftarrow \{\}$ }
    
    \While{$|S| < k$ \textbf{and} $|C| > 0$}{
    
        $s^{*} \leftarrow \argmax_{s \in C}{ROUGE(S, R)}$
        
        $S \leftarrow S \cup \{s^{*}\}$
        
        $C \leftarrow C \setminus \{s^{*}\}$
    }
    {\bf Return $S$}
    \caption{Greedy summarization oracle.}
    \label{alg:oracle}
\end{algorithm}

\begin{table*}[hbt!]
    \small
    \centering
    \resizebox{\textwidth}{!}{%
    \begin{tabular}{|ll|lllllll|}        
        \hline 
        Dataset & Method & l & k & \#articles & \#dates & \makecell{Comp. \\ ratio (sents)} & \makecell{Comp. \\ ratio (dates)} & duration \\ \hline         
        T17 & \textsc{Martschat 2018} & -0.116 & 0.381 & -0.421 & -0.586 $*$ & 0.298 & 0.57 $*$ & -0.376 \\         
        T17 & \textsc{Datewise} &-0.196 & 0.616 $ *$ & -0.354 & -0.714 $*$ & 0.319 & 0.638 $*$ & -0.577 $*$ \\ 
        T17 & \textsc{Clust} & 0.283 & 0.429 & -0.247 & -0.411 & 0.504 $*$ & 0.576 $*$ & -0.197 \\ \hline
        Crisis & \textsc{Datewise} & 0.19 & -0.3 & 0.393 & 0.096 & -0.271 & 0.147 & 0.028 \\ 
        Crisis & \textsc{Clust} & -0.087 & 0.038 & -0.184 & 0.029 & 0.061 & -0.037 & 0.013 \\ \hline
        Entities & \textsc{Martschat 2018} & -0.05 & -0.012 & -0.657 $*$ & -0.682 $*$ & 0.644 $*$ & 0.649 $*$ & -0.338 $*$ \\
        Entities & \textsc{Datewise} & -0.038 & -0.056 & -0.394 $*$ & -0.406 $*$ & 0.348 $*$ & 0.39 $*$ & -0.103 \\
            Entities & \textsc{Clust} & 0.028 & -0.044 & -0.461 $*$ & -0.501 $*$ & 0.422 $*$ & 0.515 $*$ & -0.358 $*$ \\
        \hline         
    \end{tabular}
    }
    \caption{Correlations between Task Properties and Method Performance.}
    \label{table:correlations}
\end{table*}

\FloatBarrier

\subsection{Running Time}
\label{appendix:running-time}
In Table \ref{table:running-time} we compare the running time of \textsc{Datewise} and \textsc{Martschat2018} on the \textsc{T17} and \textsc{Entities} datasets\footnote{On a machine with 16 3.70GHz Intel CPUs and 32GB memory.}. The implementations of both our methods and of \textsc{Martschat2018} make use of parallel computation to obtain pairwise similarities between sentences or documents where required. We do not parallelize our methods in any other way. We could not run \textsc{Martschat2018} on the \textsc{Crisis} dataset since it requires too much memory, which demonstrates the need for more scalable state-of-the-art methods. \textsc{Datewise} and \textsc{Clust} are considerably faster on both datasets, due to their "divide-and-conquer" nature: The summarization step is applied to only $l$ smaller portions of articles and sentences, instead of the entire set. Note that part of the time is required to run the evaluation tool to compute alignment-based ROUGE.

\begin{table}[!htb]
    \small
    \centering
    \resizebox{\columnwidth}{!}{
    \begin{tabular}{|l|l|l|}
        \hline
        Dataset & Method & \makecell{Avg. seconds \\ per topic}
        \\\hline
        \textsc{T17} & \textsc{Martschat2018} & 176 \\\hline
        \textsc{T17} & \textsc{Datewise} & 16 
        \\\hline
        \textsc{T17} & \textsc{Clust} & 15.9
        \\\hline
        \textsc{Entities} & \textsc{Martschat2018} & 106.3 \\\hline
        \textsc{Entities} & \textsc{Datewise} & 29.5 
        \\\hline
        \textsc{Entities} & \textsc{Clust} & 34.7
        \\\hline
    \end{tabular}
    }
    \caption{Running time comparison between current state-of-the-art method \textsc{Martschat2018} and the methods we implemented.}
    \label{table:running-time}
\end{table}

\subsection{Correlations between Performance and Dataset Characteristics}
\label{appendix:correlations}

Detailed results of correlations between different methods and different dataset characteristics are shown in Table \ref{table:correlations}.

\subsection{Output Examples}
\label{appendix:examples}
Table \ref{tab:examples} shows parts of timelines produced by different methods for a selection of dates that all methods have selected. The topics are taken from the \textsc{Entities} dataset. The examples demonstrate different levels of detail in describing particular events.

\begin{table*}[]
\footnotesize
\begin{tabular}{|p{2cm}|p{2cm}|p{10cm}|}
\hline
\multicolumn{3}{|c|}{\textsc{Datewise} (titles only)} \\ \hline
\textbf{Topic} & \textbf{Date} & \textbf{Summary} \\ \hline
Steve Jobs & 2009-01-14 & Apple boss Steve Jobs to take extended leave \\ \hline
Steve Jobs & 2011-08-25 & Steve Jobs resigns as Apple CEO \\ \hline \hline
Charles Taylor & 2010-08-09 & Mia Farrow contradicts Naomi Campbell in Charles Taylor trial \\ \hline
Charles Taylor & 2012-04-26 & Charles Taylor found guilty of abetting Sierra Leone war crimes \\ \hline \hline
\multicolumn{3}{|c|}{\textsc{Datewise}} \\ \hline
\textbf{Topic} & \textbf{Date} & \textbf{Summary} \\ \hline
Steve Jobs & 2009-01-14 & The boss of the Apple computer empire , Steve Jobs , today disclosed that his health problems have become " more complex " , prompting him to take extended leave from his role as chief executive until the end of June . \\ \hline
Steve Jobs & 2011-08-25 & Steve Apple made a followup statement : Apple 's Board of Directors today announced that Steve Jobs has resigned as Chief Executive Officer , and the Board has named Tim Cook , previously Apple 's Chief Operating Officer , as the company 's new CEO . \\ \hline \hline
Charles Taylor & 2010-08-09 & Campbell 's former agent Carole White and the actor Mia Farrow – both of whom were present at a dinner hosted by Nelson Mandela and attended by Taylor and Campbell – have given evidence in a court in the Hague today , some of which appeared to contradict the testimony given by the model last week . \\ \hline
Charles Taylor & 2012-04-26 & On Thursday Charles Taylor , warlord turned president of Liberia , was convicted of aiding and abetting war crimes by the Sierra Leone special court in The Hague . \\ \hline \hline
\multicolumn{3}{|c|}{\textsc{Clust}} \\ \hline
\textbf{Topic} & \textbf{Date} & \textbf{Summary} \\ \hline
Steve Jobs & 2009-01-14 & In his message to staff , Jobs said : " Unfortunately , the curiosity over my personal health continues to be a distraction not only for me and my family , but everyone else at Apple as well . \\ \hline
Steve Jobs & 2011-08-25 & Steve Jobs has resigned as chief executive of Apple . \\ \hline \hline
Charles Taylor & 2010-08-09 & Campbell said she was told by her former agent Carole White and the actor Mia Farrow that the diamonds came from Taylor , but otherwise she had no idea who sent them . \\ \hline
Charles Taylor & 2012-04-26 & Today , as they watched Taylor be convicted of aiding and abetting war crimes on all counts , they have seen justice done . \\ \hline \hline
\multicolumn{3}{|c|}{\textsc{Martschat2018}} \\ \hline
\textbf{Topic} & \textbf{Date} & \textbf{Summary} \\ \hline
Steve Jobs & 2009-01-14 & The boss of the Apple computer empire , Steve Jobs , today disclosed that his health problems have become " more complex " , prompting him to take extended leave from his role as chief executive until the end of June . \\\hline
Steve Jobs & 2011-08-25 & Steve Apple made a followup statement : Apple 's Board of Directors today announced that Steve Jobs has resigned as Chief Executive Officer , and the Board has named Tim Cook , previously Apple 's Chief Operating Officer , as the company 's new CEO . \\\hline \hline
Charles Taylor & 2010-08-09 & Farrow denies that she or White told Campbell that the diamonds had come from Taylor . \\\hline
Charles Taylor & 2012-04-26 & The first African head of state to be tried in an international court , Taylor will on Thursday hear the verdict of the Special Court for Sierra Leone in his five - year trial on charges of war crimes and crimes against humanity , including murder , rape , sexual slavery and using child soldiers . \\\hline
\end{tabular}
\caption{Partial timelines produced by different methods, for a fixed selection of and topics and dates.}
\label{tab:examples}
\end{table*}

\end{document}